\documentclass{article}

\usepackage{amsmath,amssymb,graphicx,url}
\newtheorem{proposition}{Proposition}
\newtheorem{lemma}{Lemma}
\newtheorem{theorem}{Lemma}

\title{Towards Optimal Algorithms for Multi-Player Bandits without Collision Sensing Information}
\usepackage{times}
\author{Wei Huang \thanks{Telecom SudParis, Institut Polytechnique de Paris, France} \,, Richard Combes \thanks{Universit\'e Paris-Saclay, CNRS, CentraleSup\'elec, Laboratoire des signaux et syst\`emes, France} \,and Cindy Trinh \thanks{Universit\'e Paris-Saclay, CNRS, CentraleSup\'elec, Laboratoire des signaux et syst\`emes, France}}
\usepackage[utf8]{inputenc} 
\usepackage[T1]{fontenc}    
\usepackage{hyperref}       
\usepackage{booktabs}       
\usepackage{amsfonts}       
\usepackage{nicefrac}       
\usepackage{microtype}      
\usepackage{xcolor}         
\usepackage{diagbox}
\usepackage{appendix}
\usepackage{bm}
\usepackage{bbm}
\usepackage{eqparbox}
\usepackage[utf8]{inputenc}
\usepackage[english]{babel}
\usepackage{algorithm}
\usepackage{algorithmic}

\newcommand{\indic}{\mathbbm{1}}
\newcommand{\EE}{\mathbb{E}}
\newcommand{\PP}{\mathbb{P}}
\newcommand{\alg}[1]{{$\tt  #1$}}
\newcommand{\pluseq}{\mathrel{+}=}
\newcommand{\ifthen}[2]{\STATE {\bf if} {#1} {\bf then} {#2} {\bf end if}}
\newcommand{\for}[2]{\STATE {\bf for} {#1} {\bf do} {#2} {\bf end for}}

\begin{document}
\maketitle
\begin{abstract}
 We propose a novel algorithm for multi-player multi-armed bandits without collision sensing information. Our algorithm circumvents two problems shared by all state-of-the-art algorithms: it does not need as an input a lower bound on the minimal expected reward of an arm, and its performance does not scale inversely proportionally to the minimal expected reward. We prove a theoretical regret upper bound to justify these claims. We complement our theoretical results with numerical experiments, showing that the proposed algorithm outperforms state-of-the-art.
\end{abstract}
\section{Introduction and Contribution}

\subsection{Multi-player Bandits}
We consider the multi-player multi-armed bandit problem with $K$ arms and $M < K$ players. At each time $t=1,...,T$, where $T$ is the time horizon, each player $m=1,...,M$ selects an arm denoted by $\pi_m(t) \in \{1,...,K\}$. If arm $k=1,...,K$ has been selected by strictly more than one player we say that a collision occurs on arm $k$, and we denote by 
$
\eta_{k}(t) = \indic \left\{ \sum_{m=1}^{M} \indic\{ \pi_m(t) = k \} \ge 2 \right\}
$
the corresponding indicator function. The reward received by player $m=1,...,M$ is denoted by
$
    r_m(t) = X_{\pi_m(t)}(t) [1-\eta_{\pi_m(t)}(t)]
$
where $(X_k(t))_{t \ge 1}$ is i.i.d. Bernoulli with mean $\mu_k$ for $k=1,...,K$. In other words, if a player $m$ is involved in a collision then he receives a reward of $0$, otherwise he receives a binary reward with mean $\mu_{\pi_m(t)}$.

\subsection{No Collision Information}

We consider the case where each player $m=1,...,M$ only observes its own reward $r_m(t)$ and nothing else. He does not observe the rewards obtained by other players, he does not know which arms have been selected by other players, and he cannot observe the collision indicator $\eta_{\pi_m(t)}$. This case is called "without collision sensing", and is the most arduous because when player $m$ gets a reward of $r_m(t) = 0$, he does not know whether this was caused by a collision so that $\eta_{\pi_m(t)}(t) = 1$ or whether this was caused because the reward of the chosen arm was null i.e. $X_{\pi_m(t)}(t) = 0$. Of course, when a reward of $r_m(t) = 1$ is obtained, he can however be sure that no collision has occurred. Therefore, the problem is fully decentralized, and no communication is allowed between the players.

\subsection{Regret Minimization}

 Recall that $\mu_1,...,\mu_K$ denotes the mean reward of each arm where no collision has occurred, and define $\mu_{(1)},....,\mu_{(K)}$ its version sorted in descending order, so that $\mu_{(1)} \ge \mu_{(2)} \ge ... \ge \mu_{(K)}.$ The goal for the players is to maximize the expected sum of rewards, and to do so, each player must select a distinct arm amongst the $M$ best arms to both avoid any collisions and maximize total reward. The regret (sometimes called pseudo-regret) is defined as the difference between the total expected reward obtained by an oracle that plays the optimal decision and the reward obtained by the players
$$
    R(T,\mu) = \sum_{t=1}^{T}  \sum_{m=1}^{M} \mu_{(m)} - \mathbb{E}( r_{m}(t) )
$$
Our goal is to design distributed algorithms in order to minimize the regret.
\subsection{Related Work}
Single-player multi-armed bandits were originally proposed by \cite{thompson} and asymptotically optimal strategies proposed by \cite{lairobbins}, \cite{klucb} amongst others. Building on the work of \cite{lairobbins}, \cite{anantharam} considered a generalization to the case where a single player can choose several arms, and proposed asymptotically optimal strategies. In fact, the problem considered by \cite{anantharam} can be seen as a centralized version of the distributed problem we consider here. Motivated by wireless communication and networking, more recent works have considered multi-player multi-armed bandits, \cite{liuzhao}, \cite{rosenski}, for instance as a natural model for a scenario where several players must access a wireless channel in a decentralized manner \cite{besson_multiplayer_revisited}. When collision sensing is available, several authors \cite{boursier_sicmmab}, \cite{wang} have shown that one can obtain roughly the same performance as in the centralized case by cleverly exploiting collisions. A harder problem is the case without collision sensing. The state-of-the-art known algorithms for this problem are SIC-MMAB2 \cite{boursier_sicmmab}, EC-SIC \cite{shi_ec_sic} and the algorithms of \cite{lugosi}. Our work improves on these algorithms, both in terms of performance, and in terms of practical applicability. Some generalizations and alternative settings were also considered. \cite{bistritz} and \cite{mehrabian} consider the more general heterogenous case where the mean of arms vary among players. \cite{magesh} and \cite{shi_ec3} consider a variant of the probem where collisions can yield a non-zero reward. \cite{bubeck2} consider the additional assumption that players have access to shared randomness, and study the minimax regret.

\subsection{Our Contribution}

The regret upper bounds for state-of-the-art algorithms and the prior information they require is presented in Table~\ref{table:stateoftheart} (\footnote{where $E$ in the EC-SIC regret bound is the error exponenent of the error correcting code used.}). We also present a regret lower bound which corresponds to the case with collision sensing, although we do not know if this bound is attainable without collision sensing. These algorithms suffer from two important issues: (i) They require a lower bound on the expected reward of the worst arm $\mu_{(K)}$, and cannot be used without this prior information. Since in practice this information is usually not available, this limits their practical applicability. In fact EC-SIC also requires as prior information the gap $\mu_{(M)} - \mu_{(M+1)}$ which is also typically unknown (ii) Their regret is proportional to $O(1/\mu_{(K)})$ which can be arbitrarily large. Namely $\mu_{(K)}$ may be very small, for instance, in wireless applications where arms represent channels, where $\mu_{k}$ is an increasing function of the signal-to-noise ratio on channel $k$, and where $\mu_1,...,\mu_K$ are weakly correleated due to frequency selective fading, this issue occurs. Indeed, if we assume that $\mu_1,...,\mu_K$ are drawn independently from some distribution over $[0,1]$, then $\mu_{(K)} = \min_{k=1,...,K} \mu_k$ is close to $0$ with overwhelming probability when $K$ is large.

Our main contribution is to propose an algorithm correcting issues (i) and (ii). The proposed algorithm requires no input information besides the number of arms $K$ and the time horizon $T$ and its regret is upper bounded as:
\begin{align*}
    R(T,\mu) = O\Big( \sum_{k > M} {\ln T \over \mu_{(M)} -  \mu_{(k)}}  + K^2 M \ln T  
    + K M^2 \ln \left({1 \over \mu_{(M)} - \mu_{(M+1)}}\right)^2  \ln T \Big)
\end{align*}
which does not depend on $\mu_{(K)}$. We also present numerical experiments which confirm our theoretical predictions. We believe that our work is a step towards closing the gap between the best known upper bound and lower bound, by showing that, just like in the case with collision sensing, the regret need not depend on $\mu_{(K)}$, and that is it not necessary to have prior information about $\mu_{(K)}$ either. 

\begin{table}
\begin{center}
\begin{tabular}{|c|c|}
\hline 
Algorithm & Regret Upper Bound \\ 
\hline 
SIC-MMAB2 & $O\Big(M \sum_{k > M} {\ln T \over \mu_{(M)} - \mu_{(k)}} + {K^2 M \over \mu_{(K)}} \ln T \Big)$ \\ 
EC-SIC & $O\Big( \sum_{k > M} {\ln T \over \mu_{(M)} - \mu_{(k)}} + {K M \over \mu_{(K)}} \ln T + {K M^2 \over E(\mu_{(K)})} \ln \left({1 \over \mu_{(M)} - \mu_{(M+1)}}\right)  \ln T  \Big)$ \\ 
Lugosi et al.  &  $O\Big( {K M  \over (\mu_{(M)} - \mu_{(M+1)})^2} \ln T \Big)$ (first algorithm)  \\
Lugosi et al.  &  $O\Big(  {K M  \over \mu_{(M)} - \mu_{(M+1)}} \ln T +  {K^2 M \over \mu_{(M)}} (\ln T)^2 \Big)$ (second algorithm)  \\
Our work & $O\Big( \sum_{k > M} {\ln T \over \mu_{(M)} - \mu_{(k)}}  + K^2 M \ln T + K M^2 \ln \left({1 \over \mu_{(M)} - \mu_{(M+1)}}\right)^2  \ln T \Big)$   \\ 
Lower bound & $\Omega\Big( \sum_{k > M} {\ln T \over \mu_{(M)} - \mu_{(k)}} \Big)$  \\ 
\hline 
\end{tabular} 
\vspace{1cm}
\begin{tabular}{|c|c|}
\hline 
Algorithm & Prior Information \\ 
\hline 
SIC-MMAB2 & $\mu_{(K)}$ \\ 
EC-SIC & $\mu_{(K)}$ and $\mu_{(M)} - \mu_{(M+1)}$ \\ 
Lugosi et al. & $\mu_{(M)}$ and $M$ \\ 
Our work & $\emptyset$  \\ 
\hline 
\end{tabular} 
\end{center}
\caption{State-of-the-art without collision sensing: regret upper bounds and prior information}
\label{table:stateoftheart}
\end{table}

\section{Algorithm and Analysis}

The proposed algorithm can be broken down into three steps and corresponding subroutines. Of course, since the proposed algorithm is distributed, all players $m=1,...,M$ apply the same algorithm based on their own observations. When providing pseudo-code, it is understood that each player follows the same pseudo code, in a distributed fashion. When stating our algorithms we use the notation $X \pluseq x$ to denote $X \gets X + x$, i.e. add $x$ to $X$. All additional subroutines as well as proofs are presented in the appendix.

\subsection{Step 1: Agreement on a Good Arm}

The first step of the proposed algorithm is the \alg{FindGoodArm} subroutine. The goal of this subroutine is for all players $m=1,...,M$ to agree on a "good" arm $\tilde{k}$, in the sense that $\mu_{\tilde{k}} \ge (1/8) \mu_{(1)}$, as well as to output a lower bound on $\mu_{\tilde{k}}$.

This subroutine is the cornerstone of our algorithm, and the main reason why our algorithm does not suffer from  (i) a regret scaling inversely proportionally to the reward of the first arm $O(1 / \mu_{(K)})$ (ii) requiring a non trivial lower bound of $\mu_{(K)}$ as an input parameter, while all state-of-the-art algorithms suffer from problems (i) and (ii). Indeed, once a good arm is identified, one can use this arm to exchange information between players, and arms with very low rewards have no bearing on performance and can be mostly ignored.

The \alg{FindGoodArm} subroutine is designed such that, with high probability, not only will all players agree on the same good arm, but also that they will all terminate the procedure at the same time. This is important to maintain synchronization, and achieving this synchronization is not trivial as shown next.

The \alg{FindGoodArm} subroutine is broken in successive phases, and each phase $p \ge 1$ has two sub-phases. In the first sub-phase of phase $p$, each player selects an arm uniformly at random 
$6 K 2^{p} \ln {2 \over \delta}$ times and observes the resulting rewards. For each arm $k=1,...,K$, each player computes 
$
\hat{\mu}_k^p
$
which is the ratio between the sum of rewards obtained from  arm $k$ and the number of times that arm $k$ has been selected. If $\hat{\mu}_k^p \ge 2^{1-p}$, we say that the player accepts arm $k$ at phase $p$ (i.e. he believes that $\mu_k$ is greater than $2^{-p}$) and otherwise he rejects arm $k$ at phase $p$. In the second sub-phase, the players attempt to reach agreement on one good arm. For all arms $\ell=1,...,K$, if a player accepts arm $\ell$ at phase $p$, he chooses an arm uniformly at random $K 2^{p} \ln {2 \over \delta}$ times and if he does not accept $\ell$ then he chooses arm $\ell$ for $K 2^{p} \ln {2 \over \delta}$ times. If arm $\ell$ yields at least one non-zero reward then we say that the player confirms arm $\ell$ in phase $p$.  If there exists $\tilde{k}$ such that the player confirms $\tilde{k}$, then the procedure stops and $\tilde{k}$ is output. This mechanism is designed so that, with high probability, the procedure does not exit until all players have agreed on a good arm. Indeed, if at least one of the players rejects arm $\ell$, then he will select arm $\ell$ all the time, and no non-zero reward can be obtained from $\ell$, preventing other players from confirming arm $\ell$.

To summarize the \alg{FindGoodArm} subroutine involves sampling arms uniformly at random more and more, until at least one of them has a large enough reward, and periodically exchanging information between players in order to reach agreement over one good arm, and once agreement is reached, the procedure stops. Lemma~\ref{lem:findgoodarm} provides an analysis of \alg{FindGoodArm} and Algorithm~\ref{algo:FindGoodArm} provides its pseudo-code.

\begin{lemma}\label{lem:findgoodarm}
    Consider the \alg{FindGoodArm} subroutine with input parameters $K$ and $\delta$. Then, with probability greater than $1 - C_1 K M (\ln {1 \over \mu_{(1)}}) \delta$: (i) The subroutine lasts at most $C_2 {K^2 \over \mu_{(1)}}  \ln {1 \over \delta}$ time slots causing at most $C_2 K^2 M \ln {1 \over \delta}$ regret. (ii) The output $\tilde{k}$ is a "good" arm i.e. $\mu_{\tilde{k}} \ge {1 \over 8} \mu_{(1)}$ (iii) The output $\tilde{\mu}$ is a lower bound on the expected reward of $\tilde{k}$ i.e. $\mu_{\tilde{k}} \ge \tilde{\mu}$ (vi) All players exit the subroutine at exactly the same time and output the same arm $\tilde{k}$ and estimate $\tilde{\mu}$, where $C_1,C_2$ are universal constants.
\end{lemma}

\begin{algorithm}[h]
\caption{\alg{FindGoodArm} (for player $m=1,...,M)$)}
\label{algo:FindGoodArm}
\begin{algorithmic}
\REQUIRE $K$: number of arms , $\delta$: confidence parameter
\ENSURE   $\tilde{k}$: good arm, $\tilde{\mu}$: lower bound on reward of $\tilde{k}$
\STATE $p \gets 0$, $\tilde{k} \gets -1$ {\it \# initialization} 
\WHILE{$\tilde{k} = -1$}  
    \STATE $p \pluseq 1$, $R[k],N[k] \gets 0$ for $k=1,...,K$ {\it \# current phase, rewards and number of samples}
    \STATE{\it \# sub-phase 1: explore arms uniformly at random}
    \FOR{$t = 1,...,6 K 2^{p} \ln {2 \over \delta}$} 
        \STATE Select arm $k \in \{1,\dots,K\}$ uniformly at random, observe reward $r$, $R[k] \pluseq  r$, $N[k] \pluseq 1$
    \ENDFOR
    \STATE{\it \# sub-phase 2: confirm accepted arms}
    \FOR{$\ell \gets 1,\dots,K$}
        \STATE $R'[k] \gets 0$ for $k=1,...,K$ {\it \#  rewards of samples} 
        \STATE{\it \# if arm $\ell$ was accepted sample arms uniformly}
        \IF{${R[\ell] \over N[\ell]} \ge 2^{1-p}$}
            \FOR{$t = 1,\dots, 2^{p} K \ln {2 \over \delta}$}
                \STATE Select arm $k \in \{1,\dots,K\}$ uniformly at random and observe reward $r$,  $R'[k] \pluseq r$
            \ENDFOR
            \STATE{\it \# if a non-zero reward is obtained, confirm arm $\ell$}
            \ifthen{$R'[\ell] \ge 1$}{$\tilde{k} \gets \ell$, $\tilde{\mu} \gets 2^{-p}$ {\bf break}} 
            \STATE{\it \hspace{-0.4cm} \# if arm $\ell$ was rejected only sample $\ell$}
        \ELSE
            \for{$t =1,\dots,2^{p} K \ln {2 \over \delta}$}{select arm $k=\ell$, observe reward $r$, $R'[k] \pluseq r$} 
        \ENDIF
    \ENDFOR
\ENDWHILE
\end{algorithmic}
\end{algorithm}

\subsection{Step 2: Rank Assignment}
The second part of our algorithm is the subroutine \alg{ VirtualMusicalChairs} which performs rank assignment, that is, assigning a different number to each of the players. This allows to perform orthogonalization and explore arms without collisions by using sequential hopping as done in~\cite{boursier_sicmmab}.

Our subroutine \alg{VirtualMusicalChairs} is based on the so-called Musical Chairs algorithm where players repeatedly choose an arm at random until they obtain a non-zero reward and then keep sampling this very same arm, which will also be their external rank. However, the Musical Chairs used in \cite{boursier_sicmmab} is not satisfactory since it samples from all arms $k=1,...,K$. Its performance is therefore limited by the worst arm leading to a regret which scales proportionally to $O(1 / \mu_{(K)})$, in addition to requiring a lower bound on $\mu_{(K)}$ as an input parameter.

\begin{figure}
    \centering
    \includegraphics[width=0.4\linewidth]{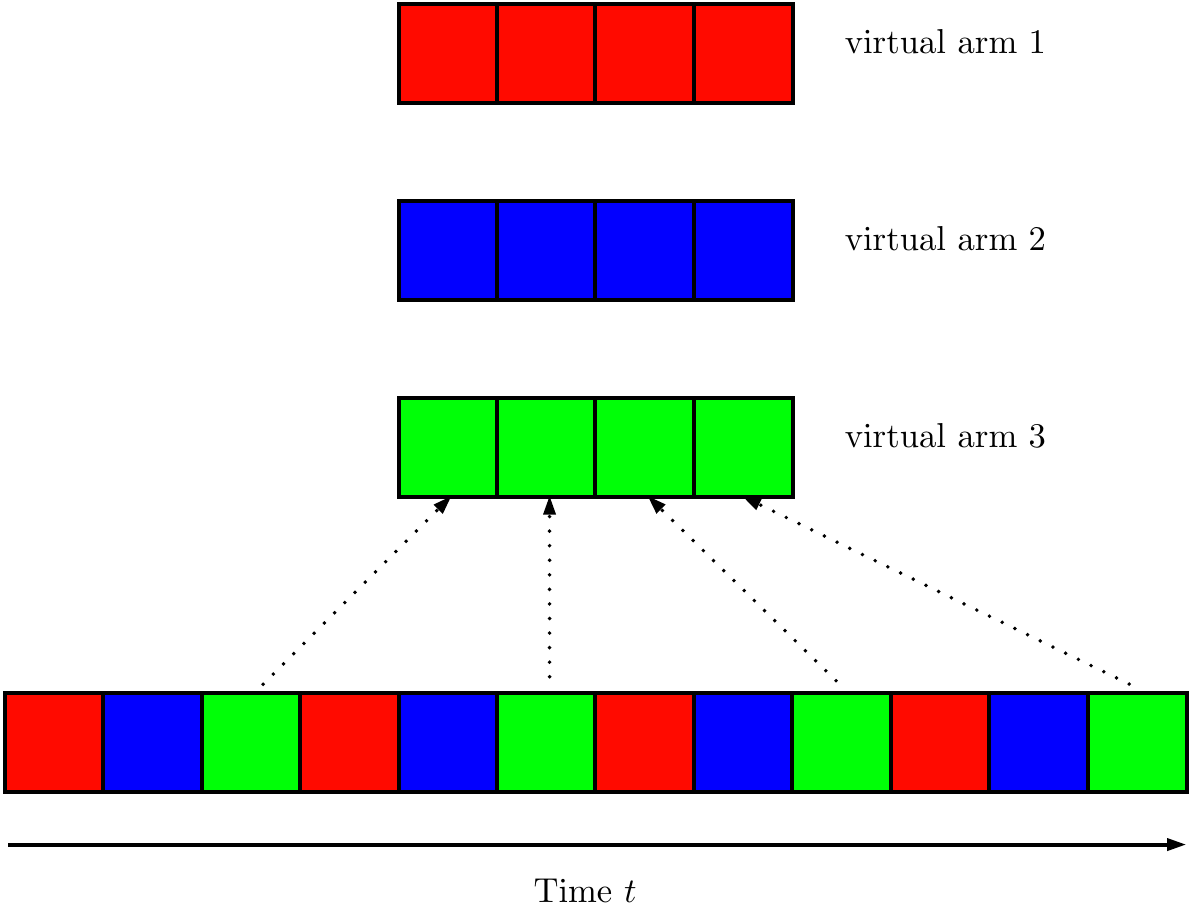}
    \caption{Mapping between a single real arm (below) and $K$ identical virtual arms (above).}
    \label{figurevirtual}
\end{figure}

Instead we propose to adapt this algorithm to only sample from the unique good arm $\tilde{k}$ derived by the \alg{FindGoodArm} subroutine. To do so we map this unique good arm into $K$ "virtual" arms by using a time division technique depicted in Figure~\ref{figurevirtual}. We divide time in $\tau$ blocks of length $K$. The $\ell$-th time slot of each block is mapped to the $\ell$-th virtual arm. At the beginning of a block, player $j$ chooses a virtual arm $\ell$ at random (without pulling it). At the $\ell$-th time slot of the block, he pulls the good arm $\tilde{k}$, and if the reward is positive, he settles on virtual arm $\ell$ until the end. Note that a player actually selects one arm every $K$ time steps on average. Note also that, due to this idea of virtual arms, \alg{VirtualMusicalChairs} can be analyzed simply by considering the classical musical chairs algorithm applied to $K$ identical arms with mean reward $\mu_{\tilde{k}}$. Lemma~\ref{lem:virtualmusicalchairs} provides an analysis of \alg{VirtualMusicalChairs} and Algorithm~\ref{algo:VirtualMusicalChairs} provides its pseudo-code.

\begin{lemma}\label{lem:virtualmusicalchairs}
    Consider $\delta > 0$. Consider the subroutine \alg{VirtualMusicalChairs} with input parameters $K$, a good arm $\tilde{k}$ and  $\tau = {K \over \mu_{\tilde{k}}} \ln {1 \over \delta}$. Then the subroutine exits after $K \tau$ time slots, incurring regret at most $8 K^2 M \ln {1 \over \delta}$ and with probability greater than $1 - C_3 K M \delta$ it assigns a distinct external rank to all of the $M$ players where $C_3$ is a universal constant. 
\end{lemma}

\begin{algorithm}
\caption{\alg{VirtualMusicalChairs} (for player $m=1,...,M$)}
\label{algo:VirtualMusicalChairs}
\begin{algorithmic}
\REQUIRE $K$: number of arms, $\tilde{k}$: a good arm, $\tau$: sampling times
\ENSURE $s$:  external rank of the player
\STATE $s \gets -1$; { \it \# rank of the player is initially unset}
\STATE{ \it \# musical chairs on the arm $\tilde{k}$}
\FOR{$t \gets 1,\dots,K \tau$}
    \STATE{ \it \# time is split in blocks of size $K$ and we select when to sample at the start of a block.}
    \IF{$\mod{(t,K)} = 1$}
        \IF{$s = -1$}
            \STATE Draw $\ell \in \{1,...,K\}$ uniformly at random { \it \# Choose a random slot if rank is unset}
        \ELSE
            \STATE $\ell \gets s$ { \it \# Choose the rank as a slot if it is set}
        \ENDIF
    \ENDIF
    \STATE{ \it \# sample the corresponding time slot}
    \IF{$\mod{(t,K)} = \ell$}
        \STATE Select arm $\tilde{k}$, and observe reward $r$
        \STATE{ \it \# set rank if it was not set yet and a non zero reward was obtained}
        \ifthen{$r > 0$ and $s=-1$}{$s \gets \ell$}
    \ELSE 
        \STATE Select an arbitrary arm in $\{1,...,K\} \setminus \{\tilde{k}\}$ 
    \ENDIF
\ENDFOR
\end{algorithmic}
\end{algorithm}

\subsection{Step 3: Computation of the Number of Players}

The third step of the proposed algorithm is the \alg{VirtualNumberPlayers} subroutine which allows players to estimate the number of players $M$. \alg{VirtualNumberPlayers} takes as input $K$ the number of arms, $\tilde{k}$ a good arm, $s$ the external rank of a player, $\tau$ a sampling time, all of which are available from the previous subroutines. 

Once again we combine the sequential hopping technique used by \cite{boursier_sicmmab} to estimate $M$ along with the idea of virtual arms, so that the procedure only samples from a good arm $\tilde{k}$, and hence is not impacted by the presence of some arms with low reward.

To estimate $M$, we initialize our estimate as $\hat{M}=1$, and perform sequential hopping for $2K$ rounds to make sure that every two different players only collide one time. Since players can be orthogonalized thanks to their external rank, a player increments $\hat{M}$ only if he gets all zero rewards during a specific period (this suggests that some collisions have occurred). As a by-product, we also output the internal rank $j \in \{1,...,M\}$, not to be confused with the external rank $s \in \{1,...,K\}$. Each player is assigned a different internal rank, and it will serve for the last step of our algorithm, in order to assign roles when exploring the various arms.

\begin{lemma}\label{lem:virtualnumberplayers}
    Consider $\delta > 0$. Consider the subroutine \alg{VirtualNumberPlayers} with input parameters $K$, a good arm $\tilde{k}$ and  $\tau = {1 \over \mu_{\tilde{k}}} \ln {1 \over \delta}$. Then the subroutine exits after $K^2 \tau$ time slots causing at most $C_4 K^2 M \ln {1 \over \delta}$ regret, 
    and with probability greater than $1 - C_5 \delta$, it both outputs a correct estimate of the number of players $\hat{M} = M$, and it assigns a distinct internal rank $j \in \{1,...,M\}$ to all of the $M$ players. Both $C_4$ and $C_5$ are universal constants. 
\end{lemma}

\begin{algorithm}
\caption{\alg{VirtualNumberPlayers} (for player $m=1,...,M$)}
\label{alg:NumberPlayers}
\begin{algorithmic}
\REQUIRE $K$: number of arms, $\tilde{k}$ a good arm, $s$: external rank of a player, $\tau$: sampling times
\ENSURE $\hat{M}$: estimated number of players, $j$: internal rank of the player
\STATE $\hat{M} \gets 1$, $\ell \gets s$, $j \gets 1$ {\it \# initialization}
\FOR{$n=1,\dots,2K$}
    \ifthen{$n > 2s$}{$\ell \gets \mod{(\ell+1,K)}$ {\it \# sequential hopping}}
    \STATE $R \gets 0$ {\it \# sum of rewards from the good arm}
    \STATE {\it \# sample from the good arm}
    \FOR{$k = 1,\dots,K$}
        \IF{$\ell \ne k$}
            \for{$t=1,\dots,\tau$}{Select an arbitrary arm in $\{1,...,K\} \setminus \{\tilde{k}\}$}
        \ELSE
            \STATE {\it \# sample from virtual arm $\tau$ times}
            \for{$t=1,\dots,\tau$}{Select arm $\tilde{k}$, observe reward $r$, $R \pluseq  r$}            \STATE {\it \# if no non-zero reward was obtained increase the estimated number of players}
            \IF{$R  = 0$}
                \STATE $\hat{M} \pluseq 1$, 
                \ifthen{$n \le 2s$}{$j \pluseq 1$}
            \ENDIF
        \ENDIF
    \ENDFOR
\ENDFOR

\end{algorithmic}
\end{algorithm}

\subsection{Step 4: Distributed Exploration and Finding the Best Arms}

The last step is the subroutine \alg{DistributedExploration}, which takes as input $K$ the number of arms, $j$ the internal rank of the player, $\tau$ the sampling times, $\hat{M}$ the estimated number of players, and outputs one arm amongst the $M$ best arms, which is assigned to the player. 

\alg{DistributedExploration} is similar to the exploration strategy used in \cite{shi_ec_sic} 
with two key improvements. First, to avoid choosing arms with low expected reward, a player sends data only through arm $\tilde{k}$ rather than through the $j$-th arm as done in \cite{shi_ec_sic}. Second, we set the quantization error to be dependent on phase p instead of the fixed quantity $\mu_{M} - \mu_{M+1}$. By doing so, the number of pulls is reduced, and the subroutine does not require this prior knowledge anymore.
Its analysis, provided in lemma~\ref{lem:distributedexploration} follows along similar lines as \cite{shi_ec_sic}, so we simply highlight the main steps of this procedure. The player with internal rank $j=1$ is called the leader, and other players are called the followers. The procedure operates in phases, and in phase $p$, the active players (the players that have not been assigned an arm yet) sample each active arm (the arms for which players have not figured out whether or not they are amongst the $M$ best arms) $2^p \lceil \ln {1 \over \delta} \rceil$ times using sequential hopping (so that no collisions occur). Then all followers send their estimates to the leader, the leader aggregates their estimates, determines which arms can be accepted (the arms for which one is sure that they are amongst the $M$ best arms) and rejected (the arms for which one is sure that they are not amongst the $M$ best arms) and then sends the sets of accepted and rejected arms back to the followers. If an arm is accepted then it is assigned to an active player, and both the arm and the player become inactive, and the player simply selects the assigned arm until the end of the procedure. If an arm is rejected then it becomes inactive. The strategy on how leader and followers act in this information exchange step is given in subroutines \alg{ComFollow} and \alg{ComLead} in appendix. In essence, the leader coordinates and takes all of the decisions, while followers simply collect samples, transmit their estimates and receive orders on what to do next. The strategy to send and receive data between leader and followers is given in subroutines \alg{EncoderSendFloat} and \alg{DecoderReceiveFloat} in appendix. 
\begin{lemma}\label{lem:distributedexploration}
    Consider $\delta > 0$ and $T \ge 0$. Consider the subroutine \alg{DistributedExploration} with input parameters $K$, $\tilde{k}$ and  $\tau = {1 \over \mu_{\tilde{k}}} \ln {1 \over \delta}$ applied with time horizon $T$ (\footnote{Applying a procedure with time horizon $T$ simply means that, if it does not terminate before $T$ on its own, then its execution is immediately stopped when time $T$ is reached.}). Then with probability greater that $1 - C_6 \delta K M (\ln T)^2 $ the subroutine incurs a regret upper bounded by
    \begin{align*}
        C_7 & \sum_{k > M} {\ln {1 \over \delta} \over \mu_{(M)} -  \mu_{(k)}} + C_{8} K M^2 \ln \left({1 \over \mu_{(M)} - \mu_{(M+1)}}\right)^2 \ln {1 \over \delta}
    \end{align*}
    and assigns one distinct arm to each player from the set of the $M$ best arms, where $C_6,C_7,C_8$ are universal constants.
\end{lemma}

\begin{algorithm}
\caption{\alg{DistributedExploration} (for players $m=1,...,M$)}
\label{algo}
\begin{algorithmic}
\REQUIRE $K$: number of arms, $j$: internal rank of a player, $\hat{M}$: number of players, $\tilde{k}$: a good arm, $\tau$: sampling times
\ENSURE $f$ an arm amongst the $M$ best arms assigned to the player
\STATE Initialize $p \gets 0$; $f \gets -1$;  {\it \# initialization}
\STATE $R[k], v[k] \gets 0$ for $k=1,...,K$ {\it \# rewards and number of samples for each arm}
\STATE {\it \# rewards and number of samples for each arm held by each players, only stored by the leader}
\IF{$j=1$}
    \for{$m=1,...,\hat{M}$ and $k=1,...,K$}{$\hat\mu[k,m],N[k,m] \gets 0$}
\ENDIF
\STATE $M' \gets \hat{M}$, ${\cal K} \gets \{1,\dots,K\}$ {\it \# number of active players and set of active arms}
\WHILE{$f = -1$}
\STATE $p \pluseq 1$ {\it \# start phase $p$}
\STATE $k \gets j$ {\it \# first sub-phase explore arms by sequential hopping}
\FOR{$t \gets 1,\dots, |\mathcal K| 2^{p} \left\lceil  \ln {1 \over \delta} \right\rceil$}
\STATE $k \gets (k + 1) \mod |\mathcal K|$ 
\STATE Select arm $k$, observe reward $r$, $R[k] \pluseq  r$, $v[k]\pluseq 1$, $E[k] \gets {R[k] \over v[k]}$ 
\ENDFOR
\STATE $Q \gets \lceil {p \over 2} + 3 \rceil$ {\it \# second sub-phase: share estimates between players}
\IF{$j = 1$}
\STATE $(f,{\cal K},M',\hat{\mu},N) \gets  $\alg{ComLeader}$(\hat{\mu},N,{\cal K},M',Q,\tau,$ $ \tilde{k},p,\delta)$ {\# player is a leader}
\ELSE 
\STATE $(f,\mathcal {\cal K},M') \gets $\alg{ComFollow}$(E,j,\mathcal {\cal K},M',Q,\tau,\tilde{k})$  {\# player is a follower}
\ENDIF
\ENDWHILE
\end{algorithmic}
\end{algorithm}

\subsection{Putting it all Together}

The complete proposed algorithm is presented in algorithm~\ref{alg:proposed}, and combines the four steps above. It is noted that our proposed algorithm is in fact a procedure to identify the set of best arms. To turn it into an algorithm for minimizing regret over a time horizon of $T$, one can simply run the proposed algorithm with confidence parameter $\delta = {1 \over T}$, and then let players select the arm that has been assigned to them by the algorithm until the time horizon runs out. The regret upper bound of Theorem~\ref{th:main} is our main result, and is simply proven by combining the four previous lemmas. As promised, unlike state-of-the-art algorithms, our algorithm does not need any prior information such as the number of arms $K$ or the reward of the worst arm $\mu_{(K)}$, and its regret does not depend on $\mu_{(K)}$. The regret bound is better than that of SIC-MMAB2 and EC-SIC, in the sense that the term proportional to $1 / \mu_{(K)}$ (which can be arbitrarily large), has been eliminated. In other words, performance is not limited by the worst arm anymore. This causes a dramatic performance gain which is seen in numerical experiments shown below.

\begin{theorem}\label{th:main}
    Consider $T \ge 0$, $\mu_{M}-\mu_{M+1}>0$. First apply the proposed algorithm with input parameters $K$ and $\delta = {1 \over T (\ln T)}$ with time horizon $T$, let $\bar{k}$ denote its output, and select arm $\bar{k}$ for the remaining time steps. Then the expected regret of this procedure is 
    \begin{align*}
        R(T) &\le C_{9} \sum_{k > M} {\ln T  \over \mu_{(M)} -  \mu_{(k)}}  + C_{10} K^2 M \ln T + C_{11} K M^2 \ln \left({1 \over \mu_{(M)} - \mu_{(M+1)}}\right)^2 \ln T
    \end{align*}
    with $C_{9},C_{10},C_{11}$ three universal constants.
\end{theorem}

\begin{algorithm}
\caption{Proposed algorithm (for player $m=1,...,M$)}
\label{alg:proposed}
\begin{algorithmic}
\REQUIRE $K$: number of arms, $\delta$: confidence level 
\ENSURE $\bar{k}$ an arm amongst the $M$ best arms assigned to the player
\STATE $(\tilde{k}, \tilde{\mu}) \gets $\alg{FindGoodArm}$(K,\delta)$  {\it \# find a good arm and a lower bound on its reward}
\STATE $s \gets $\alg{VirtualMusicalChairs}$(K, \tilde{k}, K \ln(\frac{1}{\delta})/\tilde{\mu})$  {\it \# assign external rank to each player}
\STATE $(\hat{M},j) \gets $\alg{VirtualNumberPlayers}$(K, \tilde{k}, s, \ln(\frac{1}{\delta}) / \tilde{\mu})$  {\it \# estimate the number of players and assign internal rank}
\STATE $\bar{k} \gets $\alg{DistributedExploration}$(K,j,\hat{M},\tilde{k}, \ln(\frac{1}{\delta}) / \tilde{\mu}))$ {\it \# find one arms out of the $M$ best arms}
\end{algorithmic}
\end{algorithm}

\section{Numerical Experiments}

We now compare the empirical performance of the proposed algorithm against the state-of-the-art  algorithms SIC-MMAB2 and EC-SIC. For simplicity we assume that rewards decrease from best to worst in a linear fashion $\mu_{(k)} = \mu_{(1)} + {k - 1 \over K-1} (\mu_{(K)} - \mu_{(1)})$. The regret of algorithms is averaged over $20$ (or more) independent runs, and $95\%$ confidence intervals are presented. 

\begin{figure}[H]
    \centering
    \begin{minipage}[t]{0.3\linewidth}
    \includegraphics[width=1\linewidth]{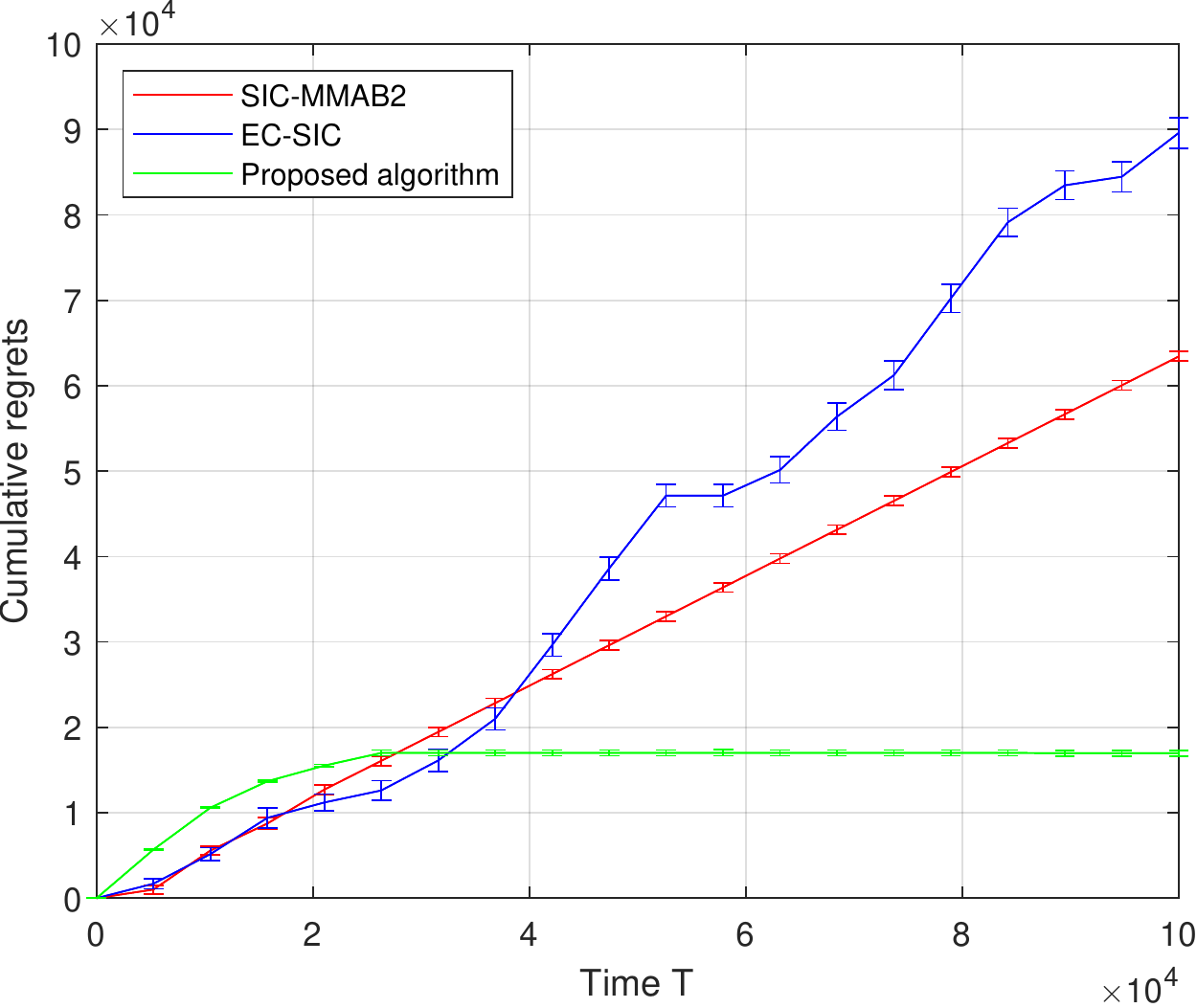}
    \caption{K=5,M=2,T=$10^5$}
    \label{figureK5M2}
    \end{minipage}
    \hspace{0.03\linewidth}
    \begin{minipage}[t]{0.3\linewidth}
    \centering
    \includegraphics[width=1\linewidth]{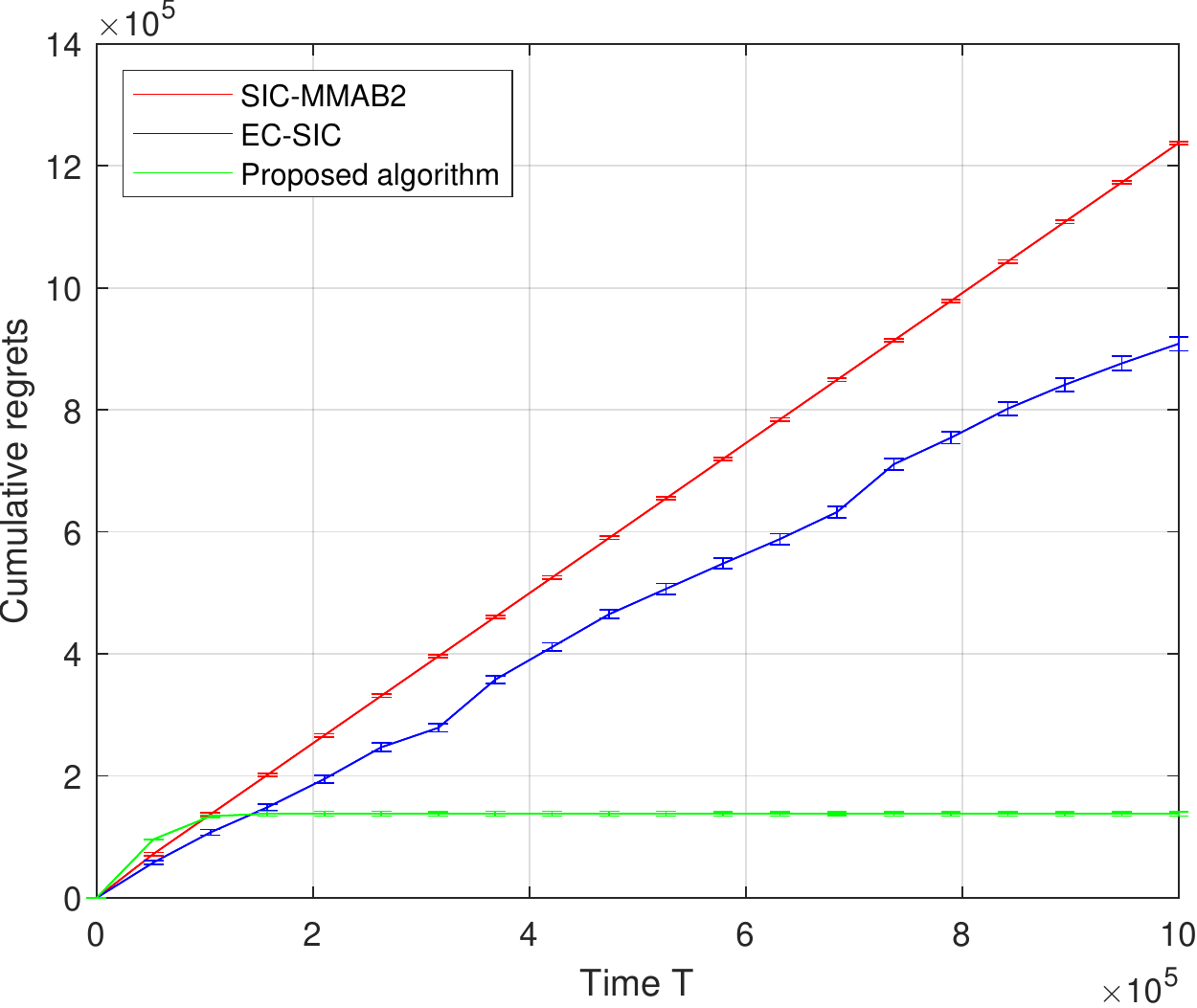}
    \caption{K=10,M=5,T=$10^6$}
    \label{figureK10M5}
    \end{minipage}
    \hspace{0.03\linewidth}
    \begin{minipage}[t]{0.3\linewidth}
    \centering
    \includegraphics[width=1\linewidth]{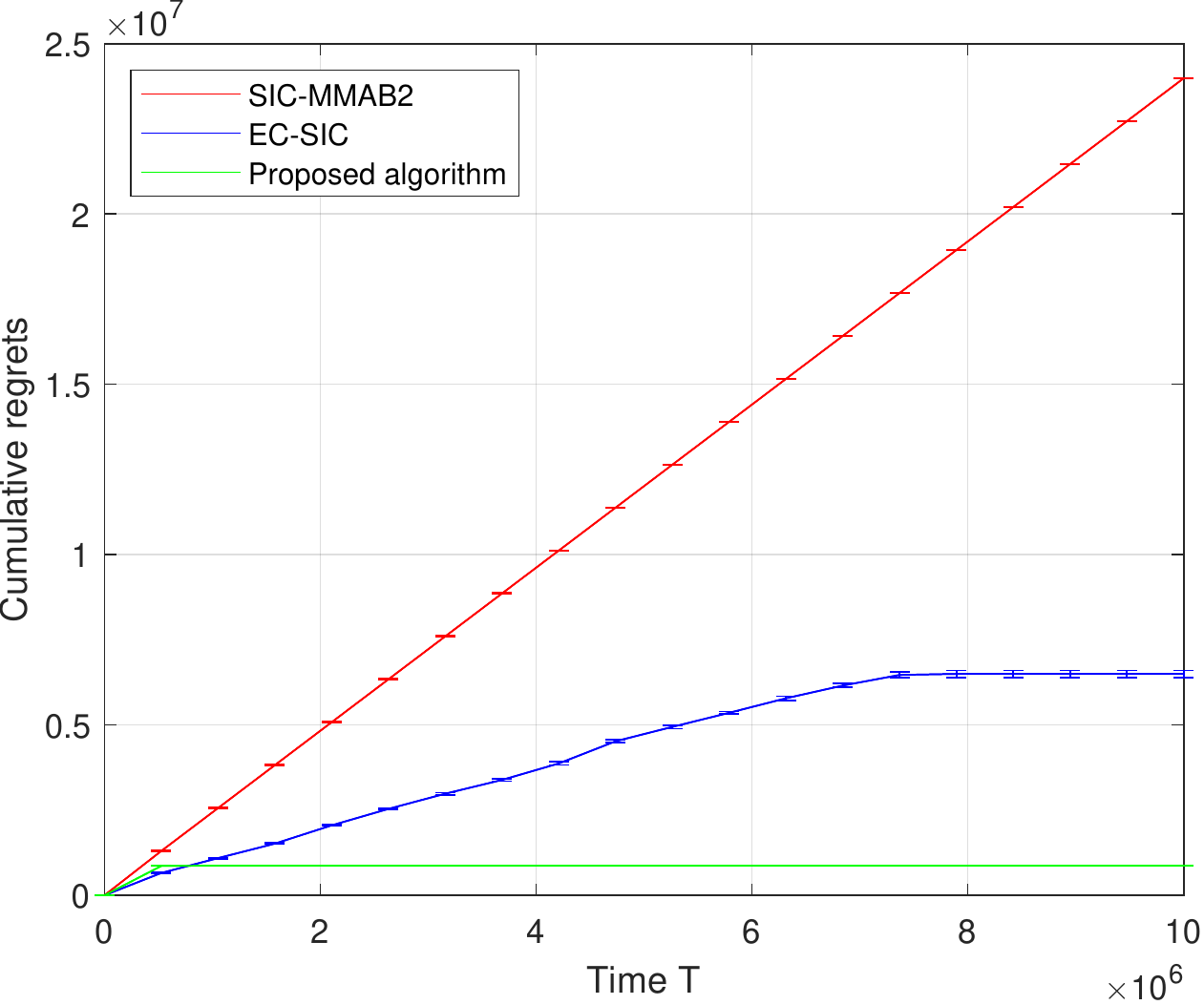}
    \caption{K=20,M=10,T=$10^7$}
    \label{figureK20M10}
    \end{minipage}
\end{figure}

{\bf Influence of the number of players} In our first set of experiments, we consider $\mu_{(1)} = 1$ and $\mu_{(K)} = 0.01$, $M = \lfloor K/2 \rfloor$. We plot the expected regret of the various algorithms for $K = 5,10,20$ in figures \ref{figureK5M2},\ref{figureK10M5} and \ref{figureK20M10} respectively. Overall, the proposed algorithm clearly outperforms EC-SIC and SIC-MMAB2, sometimes by several orders of magnitude, and the difference seems more and more severe when $K$ increases.

{\bf Influence of the gap} Figure \ref{figureK5M2} and \ref{figureK10M5} consider $\mu_{M}-\mu_{M+1} = 0.2$ and $\mu_{M}-\mu_{M+1} = 0.1$. We see that sometimes the proposed algorithm performs a bit worse than other algorithms for very small time horizons, however for larger time horizons it does outperform SIC-MMAB2 and EC-SIC which seem not to converge quickly enough.

\begin{figure}[H]
    \centering
    \begin{minipage}[t]{0.3\linewidth}
    \includegraphics[width=1\linewidth]{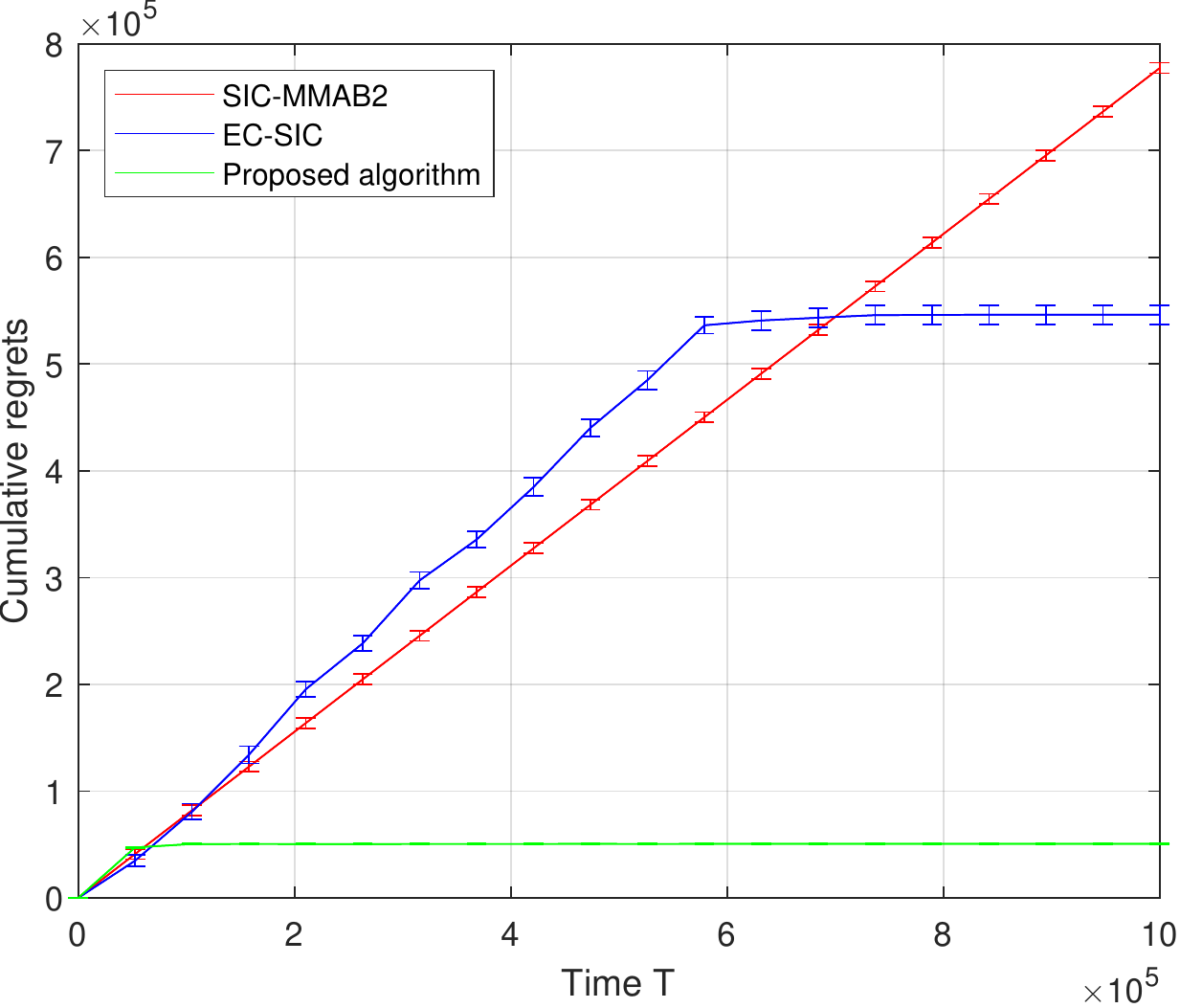}
    \caption{K=10,M=2,T=$10^6$}
    \label{figureK10M2}
    \end{minipage}
    \hspace{0.03\linewidth}
    \begin{minipage}[t]{0.3\linewidth}
    \centering
    \includegraphics[width=1\linewidth]{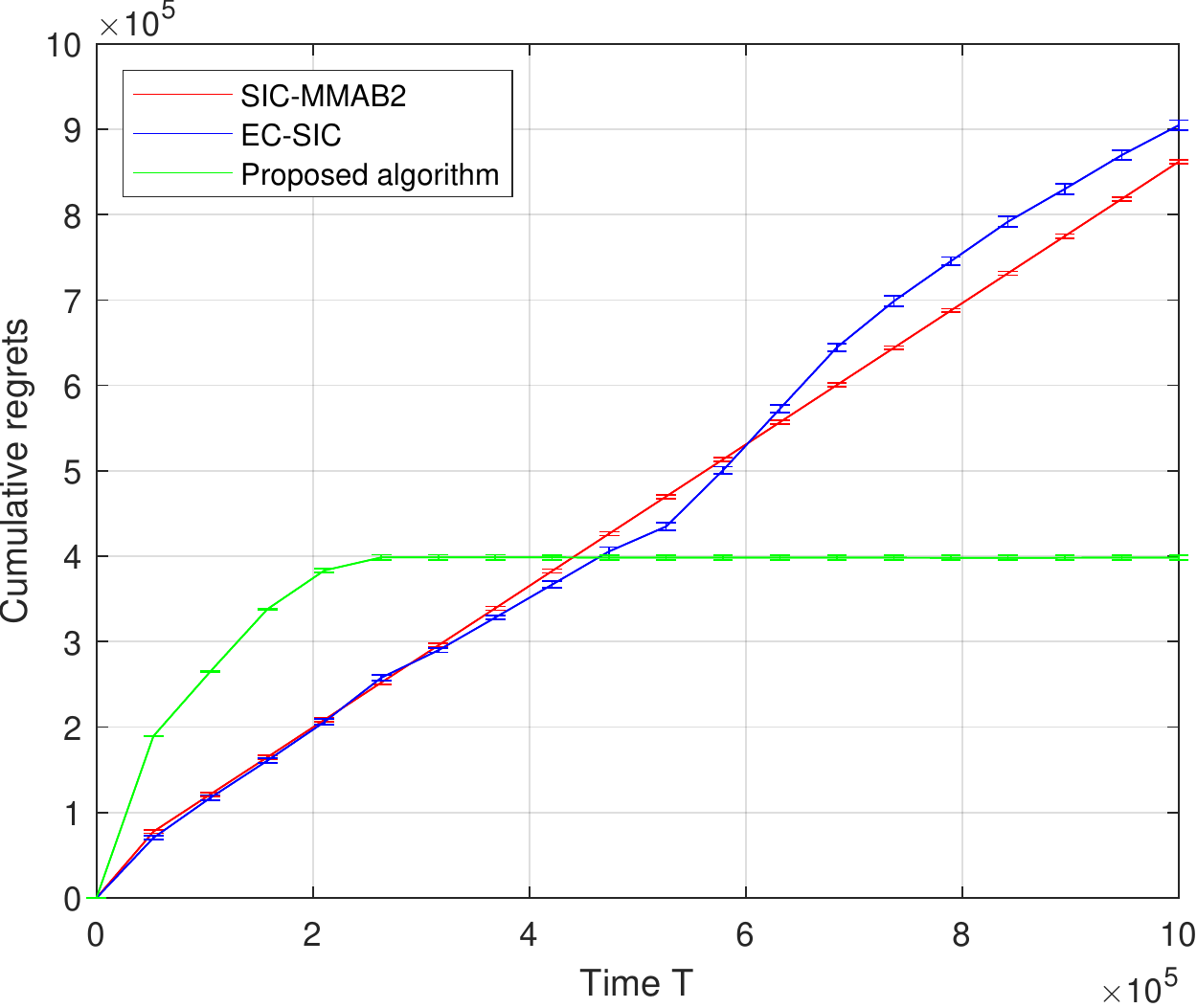}
    \caption{K=10,M=8,T=$10^6$}
    \label{figureK10M8}
    \end{minipage}
    \hspace{0.03\linewidth}
    \begin{minipage}[t]{0.3\linewidth}
    \centering
    \includegraphics[width=1\linewidth]{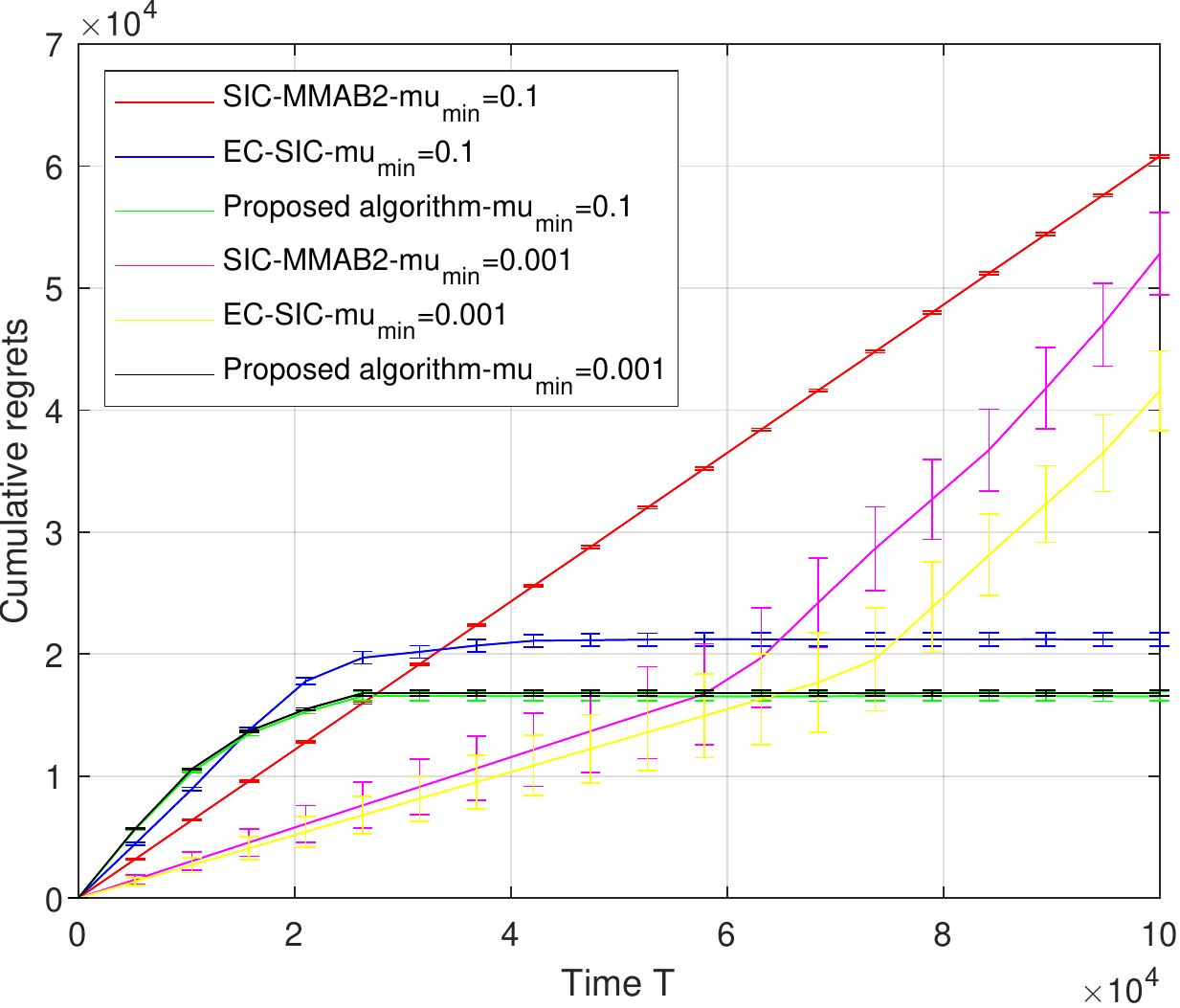}
    \caption{K=5,M=2,T=$10^5$ with different $\mu_{(K)}$}
    \label{figureK5M2mu}
    \end{minipage}
\end{figure}

{\bf Impact of the number of players} In figures \ref{figureK10M2} and \ref{figureK10M8} we compare between $M=2$ players and $M=8$ players. For $M=2$, both the proposed algorithm and EC-SIC do converge. However, when $M$ increases, the collision probability becomes larger, causing EC-SIC to spend significant time performing musical chairs to estimate $M$. 

{\bf Impact of the worst arm} We present the regret for different values of $\mu_{(K)}$ in figure \ref{figureK5M2mu}. When $\mu_{(K)}$ decreases, we can see that the performance of SIC-MMAB2 and EC-SIC is greatly affected, and for $\mu_{(K)} = 0.001$ they do not even start the exploration phase. On the other hand, no matter how small $\mu_{(K)}$ is, the cumulative regret of the proposed algorithm stays similar, which confirms our theoretical predictions.

\section{Conclusion}

In this work we have proposed a new algorithm for multi-player multi-armed bandits without collision sensing information. Through both analysis and numerical experiments, we have proven that it performs significantly better than the state-of-the-art algorithms, while at the same time requiring no input parameter other than the number of arms. The cornerstone of our approach is a novel, non trivial distributed procedure to enable players to discover a good arm without any prior information other than the number of arms. 
We believe that this result is a significant contribution to the problem as it constitutes a step towards being able to solve multi-player multi-armed bandits in a practical setting (for instance cognitive radios) where no prior information is known about the expected rewards of the arms.

{\bf Acknowledgement:} The work of Wei Huang was supported by Beyond 5G, a project of the French Government’s recovery plan ``France Relance''.

\newpage 
\bibliography{biblio}

\begin{thebibliography}{10}

\bibitem{anantharam}
Venkatachalam Anantharam, Pravin Varaiya, and Jean Walrand.
\newblock Asymptotically efficient allocation rules for the multiarmed bandit
  problem with multiple plays-part i: Iid rewards.
\newblock {\em IEEE Transactions on Automatic Control}, 32(11):968--976, 1987.

\bibitem{besson_multiplayer_revisited}
Lilian Besson and Emilie Kaufmann.
\newblock Multi-player bandits revisited.
\newblock In {\em Proc. of AISTATS}, 2018.

\bibitem{bistritz}
Ilai Bistritz and Amir Leshem.
\newblock Distributed multi-player bandits - a game of thrones approach.
\newblock In {\em Proc. of NeurIPS}, 2018.

\bibitem{boursier_sicmmab}
Etienne Boursier and Vianney Perchet.
\newblock {SIC-MMAB}: Synchronisation involves communication in multiplayer
  multi-armed bandits.
\newblock In {\em Proc. of NeurIPS}, 2019.

\bibitem{bubeck2}
S\'ebastien Bubeck, Thomas Budzinski, and Mark Sellke.
\newblock Cooperative and stochastic multi-player multi-armed bandit: Optimal
  regret with neither communication nor collisions.
\newblock In {\em Proc. of COLT}, 2021.

\bibitem{klucb}
Olivier Cappé, Aurélien Garivier, Odalric-Ambrym Maillard, Rémi Munos, and
  Gilles Stoltz.
\newblock Kullback–leibler upper confidence bounds for optimal sequential
  allocation.
\newblock {\em The Annals of Statistics}, 41(3):1516–1541, Jun 2013.

\bibitem{lairobbins}
Tze~Leung Lai and Herbert Robbins.
\newblock Asymptotically efficient adaptive allocation rules.
\newblock {\em Advances in applied mathematics}, 6(1):4--22, 1985.

\bibitem{liuzhao}
Keqin Liu and Qing Zhao.
\newblock Distributed learning in multi-armed bandit with multiple players.
\newblock {\em IEEE Transactions on Signal Processing}, 58(11):5667--5681,
  2010.

\bibitem{lugosi}
Gabor Lugosi and Abbas Mehrabian.
\newblock Multiplayer bandits without observing collision information.
\newblock arXiv 1808.08416, 2018.

\bibitem{magesh}
Akshayaa Magesh and Venugopal~V. Veeravalli.
\newblock Decentralized heterogeneous multi-player multi-armed bandits with
  non-zero rewards on collisions.
\newblock {\em IEEE Transactions on Information Theory}, 2021.

\bibitem{mehrabian}
Abbas Mehrabian, Etienne Boursier, Emilie Kaufmann, and Vianney Perchet.
\newblock A practical algorithm for multiplayer bandits when arm means vary
  among players.
\newblock In {\em Proc. of AISTATS}, 2020.

\bibitem{rosenski}
Jonathan Rosenski, Ohad Shamir, and Liran Szlak.
\newblock Multi-player bandits--a musical chairs approach.
\newblock In {\em Proc. of ICML}, 2016.

\bibitem{shi_ec3}
Chengshuai Shi and Cong Shen.
\newblock Multi-player multi-armed bandits with collision-dependent reward
  distributions.
\newblock {\em IEEE Transactions on Signal Processing}, 69:4385--4402, 2021.

\bibitem{shi_ec_sic}
Chengshuai Shi, Wei Xiong, Cong Shen, and Jing Yang.
\newblock Decentralized multi-player multi-armed bandits with no collision
  information.
\newblock In {\em Proc. of AISTATS}, 2020.

\bibitem{thompson}
William~R Thompson.
\newblock {On the likelihood that one unknown probability exceeds another in
  view of the evidence of two samples}.
\newblock {\em Biometrika}, 25(3-4):285--294, 12 1933.

\bibitem{wang}
Po-An Wang, Alexandre Proutiere, Kaito Ariu, Yassir Jedra, and Alessio Russo.
\newblock Optimal algorithms for multiplayer multi-armed bandits.
\newblock In {\em Proc. of AISTATS}, 2020.

\end{thebibliography}
\bibliographystyle{plain}
\appendix
\section{Proofs}

\subsection{Technical results}
We recall some well known concentration inequalities used for our analysis.
\begin{proposition}[Multiplicative Chernoff Bound]
    Consider $X_1,...,X_t$ i.i.d. random variables in $[0,1]$ with expectation $\mu$, then for all $\epsilon \in [0,1]$
    $$
        \PP\left( {1 \over t} \sum_{s=1}^t X_s \le (1-\epsilon)\mu \right) \le \exp\left(- {\epsilon^2 t \mu \over 2}\right) \text{ and }
        \PP\left( {1 \over t} \sum_{s=1}^t X_s \ge (1+\epsilon)\mu \right) \le \exp\left(- {\epsilon^2 t \mu \over 2+\epsilon}\right)
    $$
\end{proposition} 

\begin{proposition}[Hoeffding's Inequality]
    Consider $X_1,...,X_t$ i.i.d. random variables in $[0,1]$ with expectation $\mu$, then for all $\epsilon \in [0,1]$
    $$
        \PP\left( {1 \over t} \sum_{s=1}^t X_s \le \mu - \epsilon \right) \le \exp\left(- 2 t \epsilon^2\right)
    \text{ and }
        \PP\left( {1 \over t} \sum_{s=1}^t X_s \ge \mu + \epsilon  \right) \le \exp\left(- 2 t \epsilon^2\right)
    $$
\end{proposition} 

\subsection{Proof of Lemma 1}

When all players sample an arm uniformly at random, the probability that a player $m$ does not experience a collision is $(1 - {1 \over K})^{M-1}$, and the expected reward he obtains when selecting arm $k$ is therefore $\rho_k = (1 - {1 \over K})^{M-1} \mu_k$. 

For phase $p \ge 1$, player $m=1,...,M$ and arm $k$ we denote by $\hat{\mu}_{k,m}^p$ the empirical mean reward received by player $m$ from arm $k$ during phase $p$, denote by $t_{k,m}^p$ the corresponding number of samples. 

{\bf Acceptance in the first sub-phase} Assume that $\rho_k \ge 2^{2-p}$ then:
$$
    \PP(\hat{\mu}_{k,m}^p \le 2^{1-p}) \le \PP(\hat{\mu}_{k,m}^p \le 2^{1-p}, t_{k,m}^p \ge 3 \times 2^p \ln {2 \over \delta} ) +\PP( t_{k,m}^p \le 3 \times 2^p \ln {2 \over \delta} )
$$
Since $\hat{\mu}_{k,m}^p$ is an empirical mean of $t_{k,m}^p$ i.i.d. random variables in $[0,1]$ with expectation $\rho_k \ge 2^{2-p}$ a Chernoff bound yields:
$$
    \PP(\hat{\mu}_{k,m}^p \le 2^{1-p}, t_{k,m}^p \ge 3 \times 2^p \ln {2 \over \delta} ) \le  \exp( - {4 \over 3} \ln {2 \over \delta} ) \le {\delta \over 2}
$$
Since $t_{k,m}^p$ is an empirical mean of $6 K 2^p \ln {2 \over \delta}$ i.i.d. random variables in $[0,1]$ with expectation ${1 \over K}$ a Chernoff bound yields:
$$
    \PP( t_{k,m}^p \le 3 \times 2^p \ln {2 \over \delta} ) \le  \exp( - 3 \times 2^{p-2} \ln {2 \over \delta} ) \le {\delta \over 2}
$$
We have proven that if $\rho_k \ge 2^{2-p}$ then $\PP(\hat{\mu}_{k,m}^p \le 2^{1-p}) \le \delta$, so that arm $k$ will be accepted at phase $p$ by player $m$ with probability greater than $1 - \delta$.

{\bf Rejection in the first sub-phase} Assume that $\rho_k \le 2^{-p}$ then once again:
$$
    \PP(\hat{\mu}_{k,m}^p \ge 2^{1-p}) \le \PP(\hat{\mu}_{k,m}^p \ge 2^{1-p}, t_{k,m}^p \ge 3 \times 2^p \ln {2 \over \delta} ) + \PP( t_{k,m}^p \le 3 \times 2^p \ln {2 \over \delta} )
$$
Since $\hat{\mu}_{k,m}^p$ is an empirical mean of $t_{k,m}^p$ i.i.d. random variables in $[0,1]$ with expectation $\rho_k \le 2^{-p}$ a Chernoff bound yields:
$$
    \PP(\hat{\mu}_{k,m}^p \ge 2^{1-p}, t_{k,m}^p \ge 3 \times 2^p \ln {2 \over \delta} ) \le  \exp( - \ln {2 \over \delta} ) = {\delta \over 2}
$$
Since $t_{k,m}^p$ is an empirical mean of $6 K 2^p \ln {2 \over \delta}$ i.i.d. random variables in $[0,1]$ with expectation ${1 \over K}$ a Chernoff bound yields:
$$
    \PP( t_{k,m}^p \le 3 \times 2^p \ln {2 \over \delta} ) \le  \exp( - 3 \times 2^{p-2} \ln {2 \over \delta} ) \le {\delta \over 2}
$$
We have proven that if $\rho_k \le 2^{-p}$ then $\PP(\hat{\mu}_{k,m}^p \ge 2^{1-p}) \le \delta$, so that arm $k$ will be rejected by player $m$ at phase $p$ with probability greater than $1 - \delta$.

{\bf Confirmation in the second sub-phase} Consider arm $k$. If there exists at least one player whom did not accept this arm in the first phase then, by design, this arm cannot be confirmed. On the other hand, assume that $\rho_k \ge 2^{-p}$ and  assume that all players accept arm $k$. Denote by $R_{k,m}^p$ the total reward received by player $m$ when attempting to confirm arm $k$, which is a sum of $2^p K \ln {2 \over \delta}$ Bernoulli random variables with mean ${\rho_k \over K}$ so that:
$$
    \PP( R_{k,m}^p = 0 ) = (1 - {\rho_k \over K})^{K 2^{p} \ln {2 \over \delta}} \le \exp(- \rho_k 2^p  \ln {2 \over \delta} ) \le \exp(-  \ln {2 \over \delta} ) = {\delta \over 2}
$$
where we used the inequality $1-x \le e^{-x}$. We have proven that, if $\rho_k \ge 2^{-p}$ and that all players accept arm $k$, then arm $k$ will be confirmed by player $m$ in the second subphase with probability greater than $1 - {\delta \over 2}$ and the procedure will terminate with probability greater than $1 - {\delta \over 2}$.

{\bf Putting things together} We can now check that all statements of the theorem are true. We will use the fact that since $\rho_k = \mu_k (1-1/K)^{M-1}$ and we have $$(1 -1/e) \mu_k \le (1-1/K)^{K-1} \mu_k \le \rho_k \le \mu_k.$$ 

Statement (i) Define $k^\star \in \arg\max_{k=1,...,K} \rho_k$.  Define $p^\star$ the smallest $p$ such that $\rho_{k^\star} \ge 2^{2-p}$. With probability greater than $1 - O(M \delta)$ all players will accept arm $k^\star$ in phase $p^\star$ and with probability greater than $1 - O(M \delta)$ all players will confirm arm $k^\star$ so that the procedure will terminate. So, with probability greater than $1 - O(M \delta)$, the procedure terminates at or before phase $p^\star$. The duration of phase $p$ is $(6 + K) K 2^p \ln {1 \over \delta}$, so that the duration of the procedure up to phase $p^\star$ is:
$$
    (6 + K) K \ln {1 \over \delta} \sum_{p=1}^{p^\star} 2^p  = O(K^2 2^{p^\star} \ln {1 \over \delta}) = O( {K^2 \over \rho_{k^\star}} \ln {1 \over \delta}) = O( {K^2 \over \mu_{k^\star}} \ln {1 \over \delta})
$$
using the fact that $\rho_k \ge (1-1/e) \mu_k$ which is the announced duration. Furthermore, the amount of regret caused by any procedure is at most $M \mu_{k^\star}$ times its duration, therefore the procedure incurs regret at most
$$
    O( K^2 M \ln {1 \over \delta})
$$

Statement (ii) Denote by $\tilde{k}$ the output of the procedure. Assume that  $\rho_{\tilde{k}} \le 2^{-p^{\star}}$ then 
There must exist some player $m$ that accepts this arm at some phase $1 \le p \le p^\star$, and since $\rho_{\tilde{k}} \le 2^{-p}$, this happens with probability at most $O(\delta p^\star K M)$. On the other hand if $\rho_{\tilde{k}} \ge 2^{-p^{\star}}$ we have $\rho_{\tilde{k}} \ge  {1 \over 8} \rho_{k^\star}$ so that $\mu_{\tilde{k}} \ge  {1 \over 8}  \mu_{k^\star}$ and $\tilde{k}$ is a good arm. Hence with probability at least $1 - O(MKp^\star \delta)$ the procedure outputs a good arm.

Statement (iii) Assume that the procedure terminates at phase $1 \le p \le p^\star$. With probability at least $1 - O(MKp^\star \delta)$ we have that $\rho_{\tilde{k}} \ge 2^{-p}$ since $\tilde{k}$ must have been accepted by at least one player. Therefore $\mu_{\tilde{k}} \ge \rho_{\tilde{k}} \ge 2^{-p} = \bar{\mu}$, and the output $\bar{\mu}$ is indeed a lower bound on the expected reward of $\tilde{k}$ with probability at least $1 - O( MKp^\star \delta)$.

Statement (vi) By design of the procedure, all players must output the same arm unless they do not exit simultaneously. Denote by $\tilde{k}$ the output of the procedure. If not all players exit the procedure simultaneously, then there must exists a phase $p=1,...,p^\star$ and an arm $k$ such that all players accept arm $k$ but at least one of them does not confirm it. This happens with probability at most $O(M p^\star \delta)$, therefore, with probability at least $1 - O(M p^\star \delta)$ all players exit the procedure simultaneously. 

We also note that $p^\star = O (\ln {1 \over \mu_{k^\star}})$. This completes the proof.

\subsection{Proof of Lemmas 2 and 3}

By design \alg{VirtualMusicalChairs} and \alg{VirtualNumberPlayers} terminate after exactly $K \tau$ and $2K \tau$ time steps respectively. Furthermore applying those two procedures is equivalent to running the initialization phase of SIC-MMAB used by \cite{boursier_sicmmab}, in a system with $K$ identical arms with mean reward $\mu_{\tilde{k}}$. Therefore, from \cite{boursier_sicmmab}[Lemma 11], we have that the both of those procedures succeed with probability at least $1 - O(M K \delta)$. Also, as stated above, the amount of regret caused by any procedure is at most $M \mu_{k^\star}$ times its duration, therefore both procedures incurs regret at most
$$
    O( K \tau M  \mu_{k^\star}) = O(K M \ln {1 \over \delta} )
$$

\subsection{Proof of Lemma 4}

We now analyze the \alg{DistributedExploration} subroutine.

{\bf Maximal number of phases} It is noted that phase $p$ lasts $|{\cal K}| 2^p \lceil \ln {1 \over \delta} \rceil  \ge 2^p \lceil \ln {1 \over \delta} \rceil$. Since the time horizon is $T$, the maximal number of phases $p^\star$ must satisfy
$$
    2^{p^\star} \le 2^{p^\star+1} \ln {1 \over \delta}  =\ln {1 \over \delta} \sum_{p=1}^{p^\star} 2^p \le T
$$
therefore the number of phases is upper bounded by 
$
    p^\star \le  \ln_2 T.
$

{\bf Communication errors} Consider the subroutines \alg{EncoderSendFloat} and \alg{DecoderReceiveFloat} float, which allow to send/receive $Q$ bits each time they are called.  By design, in phase $p$ the total number of bits sent is  at most
$$
3 K M Q = 3 K M \lceil {p \over 2} + 3 \rceil
$$
Hence the total number of bits sent across phases is 
$$
    O(K M (p^{\star})^2 ) 
$$
Since, by design of the communication protocol and the fact that $ \tau = {1 \over \mu_{\tilde{k}}} \ln {1 \over \delta}$, each time a bit is sent, it is decoded correctly with probability greater than $1 - \delta$. Therefore, with probability greater than $1 - O(\delta K M (p^{\star})^2)$, there are no communication errors.

{\bf Estimation errors} Now assume that no communication errors have taken place, and denote by $\hat{\mu}^p_k$ the estimate of the mean reward of arm $k$ computed by the leader in phase $k$, $t^p_k$ the corresponding number of samples and 
$$
    B^p_k = \sqrt{(2 \ln {1 \over \delta}) / t^p_k} + 2^{-{p \over 2} - 3}
$$
We say that there is an estimation error for $k$ in phase $p$ if $| \hat{\mu}^p_k - \mu_k | \ge B^p_k$. We have:
$$
    | \hat{\mu}^p_k - \mu_k | \le | \hat{\mu}^p_k - \EE(\hat{\mu}^p_k )  | + |\mu_k - \EE(\hat{\mu}^p_k)|
$$
The first term is a random fluctuation and the second term is a quantification error due to the fact that players send quantified estimates. Since $Q = {p \over 2} + 3$ the error due to quantification is smaller than
$$
    |\mu_k - \EE(\hat{\mu}^p_k)| \le 2^{-{p \over 2} - 3}
$$
And using Hoeffding's inequality yields an upper bound for the estimation error probability:
$$
    \PP(| \hat{\mu}^p_k - \mu_k | \ge B^p_k )  \le \PP(| \hat{\mu}^p_k - \EE(\hat{\mu}^p_k )  | \ge \sqrt{(2 \ln {1 \over \delta}) /t^p_k}) \le \delta
$$
Considering all arms and all phases, we have proven that with probability at least $1 - O(\delta K p^\star)$, no estimation error occurs during a run of the algorithm. 

{\bf Behaviour on a clean run} We say that we have a clean run if no communication errors and no estimation errors occur throughout the run of the algorithm. We have already established that a clean runs occurs with probability at least $1 - O(\delta K M (p^{\star})^2)$. On a clean run, we always have that
$$
    \hat{\mu}_k^p + B_k^p \le \mu_k
$$
and
$$
   \hat{\mu}_k^p - B_k^p \ge \mu_k
$$
so that one can readily check that if $k$ is amongst the $M$ best arm, it can never be rejected, and that if $k$ is not amongst the $M$ best arms, if it is still active at phase $p$ it will be rejected if $4 B_k^p \le \mu_{(M)} - \mu_{k}$. Following the same logic as \cite{boursier_sicmmab}, one can establish that $k$ is selected at most $O({\ln {1 \over \delta} \over (\mu_{(M)} - \mu_k)^2})$ times,  and that the maximal number of phases is upper bounded by:
$$
    p^\star \le O \left( \ln \left({1 \over \mu_{(M)} - \mu_{(M+1)}}\right) \right)
$$  

{\bf Regret on a clean run} We can complete the proof by upper bounding the regret on a clean run.

We first turn to the regret caused by communication. Sending a bit using the described protocol incurs at most $M \mu_{(1)} \tau = M \ln {1 \over \delta}$ regret and $O(K M (p^{\star})^2)$ bits must be sent, therefore the regret caused by communication is upper bounded by
$$
    O\left(K M^2 \ln \left({1 \over \mu_{(M)} - \mu_{(M+1)}}\right)^2 \ln {1 \over \delta} \right)
$$
Since no collisions occur, optimal arms are never eliminated, and any suboptimal arm $k$ is selected at most $O({\ln( 1 / \delta) \over (\mu_{(M)} - \mu_k)^2})$ times, the total regret caused by exploration is
$
    O( \sum_{k > M} {\ln {1 \over \delta} \over \mu_{(M)} - \mu_k })
$.
Therefore, as announced, with probability at least than $1 - O(\delta K M (\ln T)^2)$ a clean run occurs, and the resulting regret is
$$
        O\left(K M^2 \ln \left({1 \over \mu_{(M)} - \mu_{(M+1)}}\right)^2 \ln {1 \over \delta} +  \sum_{k > M} {\ln {1 \over \delta} \over \mu_{(M)} - \mu_k}  \right) 
$$
which completes the proof.

\section{Additional Subroutines}
We provide here the pseudo-codes for the remaining subroutines. These subroutines are, we recall, described from the point of view of a single player, which is the natural way to describe a decentralized algorithm. It is noted that some subroutines are broken over several pages.  \alg{EncoderSendFloat} and \alg{DecoderReceiveFloat} allow to exchange floating point numbers, and can be modified in a straighforward manner to deal with integers, yielding \alg{EncoderSendInt} and \alg{DecoderReceiveInt}, whose pseudo-codes are omitted.

\begin{algorithm}[t]
\label{alg:communicationleader}
\caption{\alg{ComLeader}}
\begin{algorithmic}[H]
\REQUIRE 
$\hat{\mu}$: estimates held by each player, $N$: number of samples held by players, ${\cal K}$: set of active arms, $M'$: number of active players, $Q$ message length, $\tau$: sampling time, $\tilde{k}$: good arm, $p$: phase, $\delta$: confidence parameter
\ENSURE $f$ one of the $M$ best arms to be assigned to player, $\bar{{\cal K}}$ updated version set of active arms, $\bar{M}'$ updated number of active players, $\bar{\mu}$ updated estimates held by players, $\bar{N}$ updates number of samples held by players

\STATE $\bar{\mu} \gets \hat{\mu}$, $\bar{N} \gets N$
\STATE {\# \it receive the updated values of the estimates held by active players}
\FOR{$i \gets 2,\dots,M$} 
    \FOR{$k \in \mathcal K$}
        \STATE $\bar{\mu}[k,i] \gets $ \alg{DecoderReceiveFloat}$(\tilde{k}, \tau, Q)$ 
        \STATE $\bar{N}[k,i] \gets \bar{N}[k,i] + 2^{p} \left\lceil  \ln {1 \over \delta} \right\rceil$
    \ENDFOR
\ENDFOR
\STATE {\# \it compute the estimates aggregated across all players decide which arms to accept / reject}
\FOR{$k \in {\cal K}$}
    \STATE {\# \it compute the estimate aggregated across all players}
    \STATE $\rho[k] \gets (\sum_{i=1}^M \bar{\mu}[k,i] \bar{N})/(\sum_{i=1}^M \bar{N}[k,i])$
    \STATE {\# \it compute confidence radius}
    \STATE $B[k] \gets \sqrt{(2 \ln {1 \over \delta}) / (\sum_{i=1}^M \bar{N}[k,i])} + 2^{-{p \over 2} - 3}$
    \STATE {\# \it accept arm}
    \IF{$|\{i \in \mathcal K : \rho[k] - B[k] \geq  \rho[i] + B[i]\}| \geq |\mathcal K| - M'$}
        \STATE Add $k$ to $C[.,1]$
    \ENDIF    
    \STATE {\# \it reject arm}
    \IF{$|\{i \in \mathcal K : \rho[i] - B[i]  \geq  \rho[k]  + B[k] \}| \geq M'$}
        \STATE Add $k$ to $C[.,2]$
    \ENDIF
\ENDFOR
\STATE {\# \it message size to send accepted / rejected arms}
\STATE $Q' \gets \lceil \log_2 |{\cal K}|\rceil$
\STATE {\# \it send the size of the sets of accepted / rejected arms}
\FOR{$i \gets 2, \dots, M'$ and $s=1,2$}
\STATE \alg{EncoderSendInt}$(\mathcal K,\tilde{k}, \tau, Q', {\bf length}(C[.,s]))$
\ENDFOR
\STATE {\# \it send the contents of the sets of accepted / rejected arms}
\FOR{$i \gets 2, \dots, M'$ and $s=1,2$ and $k \in C[.,s]$}
\STATE \alg{EncoderSendInt}$(\mathcal K,\tilde{k},\tau,Q',k)$
\ENDFOR
\STATE $C' \gets C$ and remove $\tilde{k}$ from $C'[,.1]$
\end{algorithmic}
\end{algorithm}

\begin{algorithm}                     
\begin{algorithmic}
\IF{$M' = {\bf length}(C'[.,1])$ and $C=C'$}
    \STATE $f \gets C[M',1]$
\ELSIF{$M'-1 = {\bf length}(C'[.,1])$ and $C \ne C'$}
\STATE {\# \it assign the good arm to the leader}
    \STATE $f \gets \tilde{k}$
\ELSE
\STATE {\# \it otherwise make accepted and rejected arms inactive and update number of active players}
\STATE $\bar{M} \gets M' - {\bf length}(C[.,1])$
\STATE ${\mathcal K}' \gets {\mathcal K}$
\FOR{$k \in C'$}
    \STATE Remove $k$ from ${\cal K}'$
\ENDFOR
\ENDIF
\end{algorithmic}
\end{algorithm}

\begin{algorithm}[t]
\caption{\alg{ComFollow}}
\begin{algorithmic}[H]
\REQUIRE $E$: estimated reward of each arm by the player, $j$: rank of the player, ${\cal K}$: set of active arms, $M'$: number of active players, $Q$: message size, $\tau$: sampling time, $\tilde{k}$: a good arm
\ENSURE ,
$f$ one of the $M$ best arms to be assigned to player, $\bar{{\cal K}}$ updated version set of active arms, $\bar{M}'$ updated number of active players
\STATE ${\cal K}' \gets {\cal K} \setminus \{ \tilde{k}\}$
\STATE {\# \it send reward estimates of active arms to the leader}
\FOR{$i = 2,\dots,M'$}
\IF{$j = i$}
\FOR{$k \in {\cal K}$}
\STATE \alg{EncoderSendFloat}$({\cal K},\tilde{k},\tau,Q, E[k])$
\ENDFOR
\ELSE
\FOR{$t=1,..., |{\cal K}| \tau Q$}
    \STATE Select the $(j \mod |{\cal K}'|)$-th  arm in set ${\cal K}'$
\ENDFOR
\ENDIF
\ENDFOR
\STATE {\# \it message size to send accepted / rejected arms}
\STATE $Q' \gets \lceil \log_2 |{\cal K}|\rceil$
\STATE {\# \it receive the sizes of the sets of accepted and rejected arms}
\FOR{$i \gets 2, \dots, M'$}
\IF{$j = i$}
\FOR{$s=1,2$}
    \STATE $N[s] \gets$ \alg{DecoderReceiveInt}$(\tilde{k}, \tau, Q'$)
\ENDFOR
\ELSE
\FOR{$t=1,...,2 \tau Q'$}
    \STATE Select the $(j \mod |{\cal K}'|)$-th  arm in set ${\cal K}'$
\ENDFOR
\ENDIF
\ENDFOR
\STATE {\# \it receive the contents of the sets of accepted and rejected arms}
\FOR{$i \gets 2, \dots, M'$}
\IF{$j = i$}
\FOR{$s=1,2$ and $q=1,...,N[s]$}
        \STATE $C[q,s] \gets$ \alg{DecoderReceiveInt} $(\tilde{k},\tau,Q')$ 
\ENDFOR
\ELSE
\FOR{$t=1,...,\tau Q'(N[1]+N[2])$}
    \STATE Select the $(j \mod |{\cal K}'|)$-th  arm in set ${\cal K}'$
\ENDFOR
\ENDIF
\ENDFOR
\end{algorithmic}
\end{algorithm}

\begin{algorithm}                     
\begin{algorithmic}

\STATE {\# \it update the set of active arms, and the active players}
\STATE {\# \it avoid assigning the good arm to followers}
\STATE $C \gets C'$
\STATE Remove $\tilde{k}$ from $C'[.,1]$
\STATE {\# \it if an accepted arm can be assigned to player then do so}
\IF{$M' - j + 1 \leq {\bf length}(C'[.,1])$}
\STATE $f \gets C'[M'-j+1,1]$
\STATE {\# \it otherwise make accepted and rejected arms inactive and update number of active players}
\ELSE
\STATE $\bar{M} \gets M' - {\bf length}(C'[.,1])$
\STATE $\bar{\mathcal K} \gets {\mathcal K}$
\FOR{$k \in C'$}
    \STATE Remove $k$ from $\bar{\mathcal K}$
\ENDFOR
\ENDIF
\label{alg:communicationfollow}
\end{algorithmic}
\end{algorithm}

\begin{algorithm}[t]
\caption{\alg{EncoderSendFloat}}
\begin{algorithmic}[H]
\REQUIRE ${\mathcal K}$: a subset of arms, $\tilde{k}$: a good arm, $\tau$: a sampling time, $Q$: the message size, $\mu \in [0,1]$ a real number to send
\STATE {\# \it convert the number to send to a binary message of size $Q$}
\STATE $S \gets $\alg{FloatToBinary}$(\mu,Q)$ 
\STATE {\# \it send each bit of the binary message}
\FOR{$q=1,\dots, Q$}
    \IF{$S[q] = 1$}
		\STATE {\# \it send a $1$ bit}        
        \STATE $\ell \gets q \mod |\mathcal K \backslash \tilde{k}|$
        \STATE $k \gets$ the $\ell$-th arm in set ${\cal K} \backslash \tilde{k}$ 
    \ELSE
        \STATE {\# \it send a $0$ bit}
        \STATE $k \gets \tilde{k}$
    \ENDIF
    \FOR{$t=1,\dots,\tau$}
        \STATE Select arm $k$ 
    \ENDFOR
\ENDFOR
\label{alg:encodersendfloat}
\end{algorithmic}
\end{algorithm}

\begin{algorithm}[t]
\caption{\alg{DecoderReceiveFloat}}
\begin{algorithmic}[H]
\REQUIRE $\tilde{k}$: a good arm, $\tau$: a sampling time, $Q$: the message size
\ENSURE $\mu \in [0,1]$ a received real number to send
\STATE {\# \it decode each bit of the binary message}
\FOR{$q=1,\dots, Q$}
    \STATE $B[q] \gets 0$
    \FOR{$t=1,\dots,\tau$}
        \STATE Select arm $\tilde{k}$ and observe reward $r$
        \STATE {\# \it decode a $1$ bit if at least one non zero reward is obtained}
        \IF{$r > 0$}
            \STATE $B[q] \gets 1$
        \ENDIF
    \ENDFOR
\ENDFOR
\STATE {\# \it convert the received binary message to a real number}
\STATE $\mu \gets $\alg{BinaryToFloat}$(B,Q)$
\label{alg:decoderreceivefloat}
\end{algorithmic}
\end{algorithm}

\end{document}